# Too much evidence, too little time: From text to actionable recommendations through multi-objective evidence reasoning

Adela BÂRA, Simona-Vasilica OPREA*
*Department of Economic Informatics and Cybernetics, Bucharest University of Economic Studies, Bucharest, Romania*
*Corresponding author. E-mail: *simona.oprea@csie.ase.ro*

**Abstract** Evidence-based clinical decision making requires specialists to identify, evaluate and synthesize relevant scientific literature. However, PubMed searches for complex clinical cases often return hundreds of publications that cannot be reviewed manually under time constraints. This study proposes SCEPTER (Single-Case Evidence-driven PubMed-To-rEcommendation Reasoner), a framework for transforming clinical case descriptions into evidence-based recommendations. SCEPTER combines PubMed retrieval, PubMedBERT semantic ranking, large language model (LLM)-based claim extraction, evidence-level weighting, contradiction detection, consensus analysis and multi-objective Pareto claim selection. The framework generates structured evidence syntheses and grounded actionable recommendations. A Paper Q&A module further enables interactive exploration of selected publications. The proposed framework introduces multi-objective reasoning model that integrates literature support, contradiction analysis and interactive literature interrogation into a unified clinical decision-support pipeline. Evaluation on 150 case studies demonstrated that the framework reduced an average search space of 576 papers to 53 retained papers, 7 Pareto-optimal claims and 3 final recommendations, corresponding to an overall compression ratio of 192:1. Despite this reduction, the retained evidence maintained high diversity (entropy=0.901). The ablation study showed that Pareto-based selection increased evidence diversity and recommendation utility compared with conventional ranking approaches.

## 1. Introduction

Evidence-based decision making requires practitioners to integrate case-specific information with an ever-growing body of scientific literature. However, a PubMed search for a complex clinical case may return hundreds or even thousands of potentially relevant publications [1], [2]. Under time pressure, specialists cannot realistically screen, compare, reconcile conflicting findings and synthesize this volume of evidence into actionable recommendations [3]. The problem is particularly pronounced in domains such as mental health, where multiple interacting biological, psychological and behavioral factors often lead to competing explanations and heterogeneous evidence.

Recent advances in large language models (LLMs) have stimulated the development of retrieval-augmented clinical decision-support systems capable of summarizing biomedical literature and generating recommendations [4], [5], [6]. Nevertheless, several limitations remain: i) most systems rely on simple relevance ranking and treat retrieved papers as independent information sources, without explicitly modelling evidence quality or study-design hierarchy; ii) contradictory findings are often hidden within summaries rather than being surfaced and analyzed as clinically meaningful signals; iii) recommendations are frequently generated from large collections of retrieved documents without a principled mechanism for selecting the most informative evidence; iv) many systems provide limited transparency, making it difficult for practitioners to trace recommendations back to the supporting scientific literature [7], [8]. These limitations motivate the following research questions:

*RQ1:* How can large volumes of biomedical literature be transformed into a compact set of clinically relevant and actionable recommendations?

*RQ2:* How can evidence quality, relevance and literature support be jointly considered when selecting the most informative findings from retrieved studies?

*RQ3:* How can contradictory evidence be identified, quantified and communicated to support more transparent clinical decision making?

*RQ4:* How can recommendation generation remain fully traceable to the underlying scientific evidence while enabling specialists to interrogate the supporting literature using natural language?

To address these questions, we propose *SCEPTER (Single-Case Evidence-driven PubMed-To-rEcommendation Reasoner)*, a framework that transforms free-text clinical case descriptions into ranked, explainable and evidence-based recommendations. SCEPTER combines PubMed retrieval, PubMedBERT semantic ranking, LLM-based claim extraction, evidence-level weighting, contradiction detection, consensus analysis and multi-objective Pareto selection to identify the most informative claims from the retrieved literature. The framework generates structured evidence syntheses and recommendation sets while preserving provenance links to the supporting publications. In addition, an integrated Paper Q&A module enables specialists to interactively query selected scientific articles and obtain case-specific answers grounded in the retrieved evidence.

The main contributions of this paper are:

1. A complete evidence-to-recommendation pipeline that transforms free-text clinical cases into traceable recommendations.
2. A multi-objective claims selection mechanism based on Pareto optimality that jointly considers case relevance, methodological quality and literature support.
3. An explicit contradiction-analysis framework that identifies and reports conflicting findings rather than suppressing them during evidence aggregation.
4. A provenance-aware evidence exploration mechanism that links every recommendation to the underlying scientific publications and extracted claims while enabling interactive Q&A over the supporting literature.

## 2. Literature review

### 2.1 Evidence-based clinical decision making and information overload

Reviews of healthcare analytics highlight the increasing importance of data-driven decision making while emphasizing difficulties associated with information volume, heterogeneity and knowledge extraction from large datasets [9], [10]. Similar challenges have been observed in healthcare policy research, where process-tracing studies stress the need for systematic approaches capable of linking evidence, causal mechanisms and recommendations in complex clinical and organizational settings [11].

The volume of unstructured clinical information further increases cognitive burden. Visual analytics approaches have been proposed to facilitate navigation of extensive patient records and support time-sensitive decision making by extracting and organizing clinically relevant information from large document collections [12]. Likewise, research on opioid use disorder detection demonstrated the value of transformer-based models for extracting actionable insights from unstructured electronic health records to improve clinical decisions [13].

### 2.2 Clinical decision support systems and recommendation frameworks

Clinical Decision Support Systems (CDSSs) have emerged as an important mechanism for transforming large amounts of medical information into actionable recommendations. Knowledge-based CDSS architectures have demonstrated effectiveness in pediatric emergency telephone triage through ontology-driven reasoning and explainable recommendations [14]. Recent advances have extended CDSS capabilities through multi-agent architectures and LLMs, enabling support for triage, diagnosis, treatment planning, resource allocation and emergency department management [15].

Several studies have explored the use of LLMs directly in clinical decision-making tasks. GPT-4 has shown promising performance in emergency triage scenarios, although concerns remain regarding calibration and severity estimation [16]. Similar findings were reported for emergency medical service prioritization, where LLM-based assessments achieved agreement with expert paramedics [17]. In psychiatry, LLMs demonstrated strong capabilities for evidence-based diagnosis and treatment recommendation but were limited in capturing dynamic and phenomenological aspects of clinical reasoning [18]. Beyond acute care, AI-driven frameworks combining radiomics, deep learning and LLM agents have been successfully applied to generate personalized treatment recommendations in hepatocellular carcinoma, highlighting the potential of multi-agent clinical reasoning systems for precision medicine [19]. Broader perspectives on oncology further emphasize the role of generative AI and LLMs in integrating multimodal clinical evidence to support adaptive treatment strategies and complex therapeutic decisions [20].

### 2.3 Natural language processing (NLP) and semantic understanding of case narratives

Transforming unstructured case descriptions into structured knowledge has become a central research topic across multiple domains. Reviews of occupational injury analysis have shown that NLP techniques enable the extraction of causal factors and support automated decision-making systems from textual incident reports [21]. Similar approaches have been adopted in legal informatics, where BERT-based recommendation systems infer relevant legal provisions directly from case descriptions and generate recommendations without requiring domain-specific queries from users [22]. Outside healthcare, semantic matching techniques have been used to connect resumes with job descriptions, identify skill gaps and generate personalized recommendations [23], [24]. These studies demonstrate the ability of transformer-based architectures to convert complex narratives into structured representations that support decision-making processes.

### 2.4 LLMs for evidence synthesis and decision support

LLM-enabled agents have been shown to automate information gathering, document synthesis and decision support in business environments characterized by large volumes of textual communication [25]. Similar trends have been reported in digital manufacturing, finance, cybersecurity, governance and enterprise data management, where LLMs facilitate knowledge extraction, reasoning and recommendation generation from large-scale heterogeneous datasets [26], [27], [28], [29].

Despite their strong reasoning capabilities, LLMs remain limited by static training knowledge and susceptibility to hallucinations. Retrieval-Augmented Generation (RAG) has therefore become an important paradigm for grounding model outputs in external evidence. Recent studies demonstrate that RAG-enhanced systems improve recommendation accuracy, preference inference and robustness by integrating retrieved information with LLM reasoning [30]. Structured RAG architectures further improve retrieval efficiency, reduce redundancy and enhance response diversity in large-scale analytical environments [31].

### 2.5 Multi-objective reasoning and evidence prioritization

While retrieval and summarization technologies have advanced, selecting the most relevant evidence remains a challenge. Traditional ranking approaches frequently prioritize relevance scores alone, potentially overlooking evidence diversity, contradictions and uncertainty. Studies across healthcare, analytics and decision sciences emphasize that decision quality depends not only on data availability but also on the ability to balance competing objectives and integrate heterogeneous evidence sources [32], [33].

Recent research highlights the importance of explainability, uncertainty quantification, contradiction management and human oversight in AI-supported decision systems [27], [29]. Furthermore, clinical guidance frameworks advocate transparent evidence evaluation processes that consider evidence strength, conflicting findings and consensus formation before generating recommendations [34]. However, existing approaches rarely integrate evidence retrieval, semantic ranking, contradiction analysis, consensus detection, recommendation generation and interactive literature exploration within a single framework.

### 2.6 Research gap and contribution

Existing CDSSs primarily focus on specific tasks such as triage, diagnosis, treatment planning, or patient prioritization, rather than supporting the complete evidence-based decision-making process from clinical case description to actionable recommendation. While transformer models and LLMs have demonstrated strong capabilities in extracting information from unstructured clinical narratives and supporting reasoning tasks, they generally lack systematic mechanisms for evaluating evidence quality and reliability. Also, RAG-based and recommendation systems improve retrieval and summarization performance, yet they typically rank information based on relevance alone, without explicitly considering contradictory findings, consensus levels, or evidence hierarchies. Further, existing recommendation frameworks across healthcare and other domains often generate recommendations from retrieved information, providing limited transparency regarding how competing evidence is reconciled.

SCEPTER addresses these limitations providing an evidence-to-recommendation pipeline. The framework combines PubMed retrieval, PubMedBERT semantic ranking LLM-based claim extraction, evidence-level weighting, contradiction detection, consensus analysis and multi-objective Pareto optimization. The framework further provides transparent evidence synthesis and an interactive Paper Q&A module that enables users to interrogate the underlying literature.

## 3. Methodology

SCEPTER transforms a free-text clinical case into evidence-based recommendations through a sequence of retrieval, reasoning and synthesis steps. The framework begins by converting the clinician’s narrative into a structured representation and generating competing clinical hypotheses. They are used to retrieve and rank relevant literature from PubMed, after which the retrieved evidence is filtered, decomposed into atomic claims and weighted according to methodological quality. The resulting evidence pool is then consolidated through deduplication, contradiction detection, Pareto-based claim selection and consensus analysis. The curated evidence is synthesized into a clinician-readable narrative and translated into actionable recommendations, with every output traceable to its supporting publications. The framework in Figure 1 follows a *Retrieve→Score→Deduplicate&Contradiction Detection→Pareto Select→Interpret Evidence* workflow.

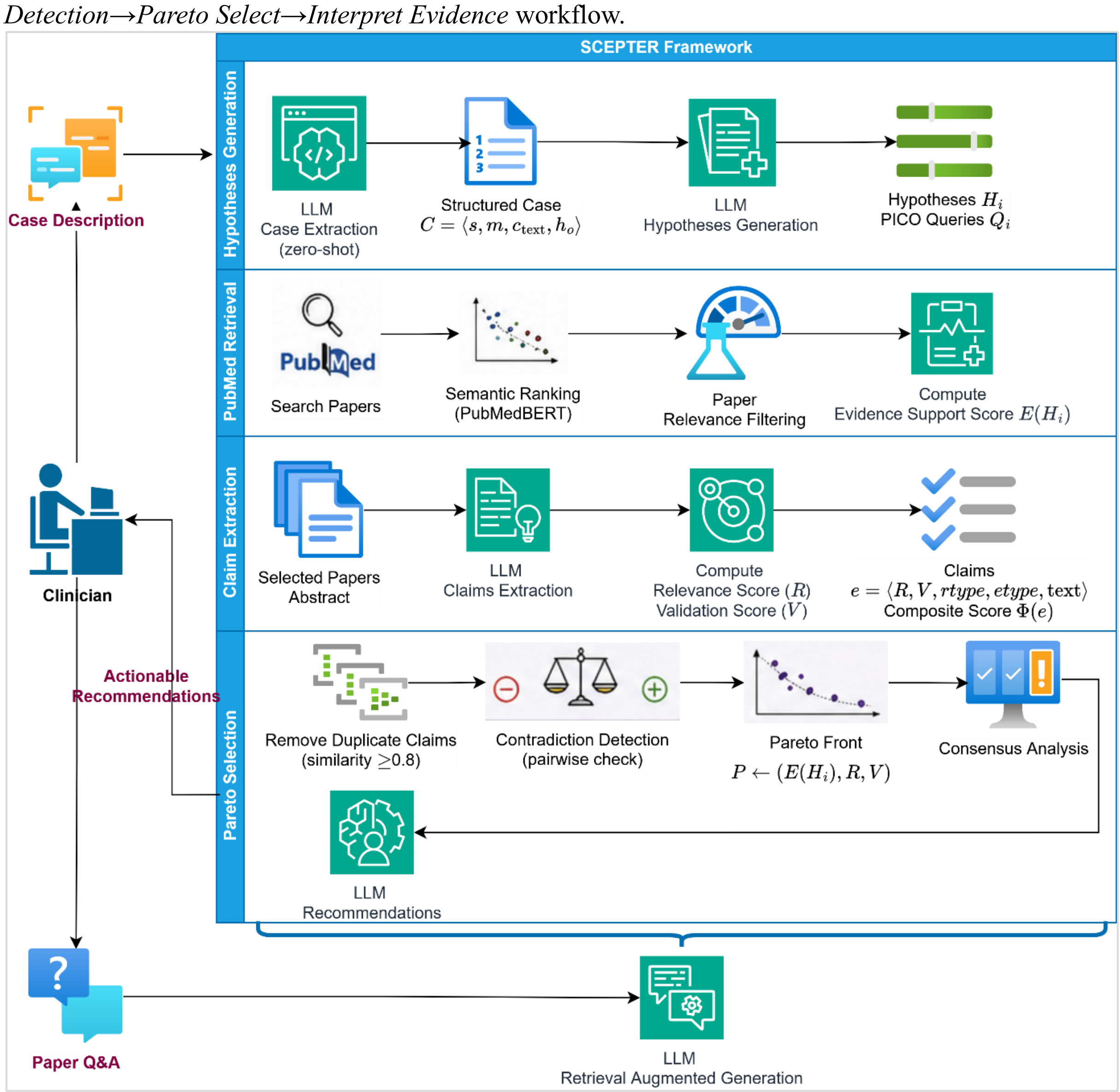

Figure 1. SCEPTER architecture

### 3.1 Case structuring

The clinical specialist inputs a free-text narrative $C_{\text{raw}}$ describing the patient's presentation, history, current medications and any initial diagnostic impressions. As downstream stages require discrete, queryable entities, the first stage transforms this narrative into a structured representation using a zero-shot LLM extraction prompt. Rather than relying on pattern matching or rule-based NLP which would require domain-specific ontology alignment, the LLM is asked to reason about clinical semantics directly from the text, which generalizes better to rare conditions and non-standard terminology.

The output is a structured case object obtained by the LLM transformation function:

$$f_{\mathcal{L}}(C_{\text{raw}}) \to C = \langle s, m, c_{text}, h_0 \rangle \tag{1}$$

where $s$ is the symptom set, $m$ the medication list, $c_{text}$ the clinical context string (including patient demographics, comorbidities and relevant history) and $h_0$ the specialist's initial hypothesis.

The LLM is prompted with a target JSON schema and instructed to be conservative. It should only populate fields that are explicitly supported by the text rather than inferring plausible values.

### 3.2 Hypothesis generation and query builder

With the structured case $C$ available, this step uses the LLM to generate a set of competing clinical hypotheses. The goal is to produce hypotheses that are *diverse,* covering different mechanistic pathways and intervention levels, rather than simply enumerating the most probable diagnoses, which an experienced clinician would already hold. Diversity is encouraged by prompting the LLM to generate three complementary hypothesis types: i) *mechanistic* that focus on potential causal processes linking factor A to outcome B (e.g., increased caffeine consumption-A exacerbates anxiety symptoms and sleep disturbances-B), ii) *intervention* examines whether a treatment or management strategy A improves a clinical outcome B (e.g., cognitive behavioral therapy for insomnia-A reduces anxiety severity-B); iii) *diagnostic* investigates whether an alternative condition A may better explain the observed symptoms B (e.g., adult ADHD-A explains chronic procrastination, concentration difficulties and academic anxiety-B). The diversification strategy directs PubMed queries toward different study designs that are most informative for each relationship type.

The LLM generates $N_H$ hypotheses $\mathcal{H} = \{H_1, \dots, H_{N_H}\}$, each hypothesis $H_i$ typed as one of $\{$*mechanistic, intervention, diagnostic*$\}$. For each hypothesis, the LLM also generates $N_Q$ queries $Q_i = \{q_{i,1}, \dots, q_{i,N_Q}\}$, each structured as $\langle P, I, C, O, focus \rangle$, where $P$-the patient population, $I$-the intervention or exposure, $C$-the comparator, $O$-the outcome and *focus* is a free-text description of what each query prioritizes.

$$f_{\mathcal{L}}(C) \rightarrow \{H_i, Q_i\} \tag{2}$$

Using multiple complementary PICO formulations per hypothesis compensates for the well-known sensitivity of PubMed keyword searches to vocabulary variation and MeSH term coverage gaps. Each variant targets a different facet of the hypothesis, for instance, one query may focus on the pharmacological mechanism while another targets the patient sub-population.

**3.3 PubMed retrieval and relevance ranking**

Evidence retrieval is a critical bottleneck in evidence-based clinical decision support. A query that is too narrow returns too few papers, while a query that is too broad introduces noise that degrades claim quality. SCEPTER addresses this tension through a four-step retrieval strategy: *i)* keyword-based PubMed search with year-range filtering; *ii)* corpus-level semantic re-ranking; *iii)* filtering by a hard relevance threshold gate; *iv)* abstract screening using LLM.

*3.3.1 Keyword-based PubMed search*

For each PICO query $q_{i,j}$, the raw query string is bounded by a publication year filter to exclude outdated evidence that may conflict with current clinical guidelines:

$$q'_{i,j} = q_{i,j} \;\wedge\; (\text{year} \in [Y_{\min}, Y_{\text{now}}]) \tag{3}$$

As multiple PICO queries per hypothesis often target overlapping bodies of literature, PMIDs are de-duplicated at the hypothesis level before full-record fetching. This reduces redundant API calls and prevents the same paper from being scored multiple times under the same hypothesis. Full records (title, abstract, authors, journal, year, MeSH[1] keywords, DOI) are retrieved through the NCBI Entrez API using the MEDLINE record format, which provides a richer set of bibliographic metadata required by the downstream stages.

*3.3.2 Semantic relevance scoring*

PubMed keyword search ranks the results by recency and MeSH term matching, not by semantic alignment with the specific clinical case. A paper that matches the query keywords but discusses a different patient population or outcome may occupy a top PubMed rank while a highly relevant paper buried further down is discarded. To correct it, we apply a semantic re-ranking step that scores each retrieved paper against a composite query representation of the actual case. Thus, a query document $d_q^{(i)}$ is constructed by concatenating all structured fields from $C$ and all PICO query variants in $Q_i$. This ensures that the query encoding captures the full clinical context rather than just the PubMed keyword string:

$$d_q^{(i)} = C \oplus \bigoplus_{q \in Q_i} q \tag{4}$$

Also, each paper $p_k$ is represented as $p_k = \{\text{title}_k \parallel \text{abstract}_k\}$. Semantic similarity is computed using *PubMedBERT* dense embeddings via the *NeuML/pubmedbert-base-embeddings sentence-transformer* model, a variant of *PubMedBERT* fine-tuned for biomedical semantic textual similarity. Both $d_q$ and all $p_k$ are encoded in a single batched forward pass, producing L2-normalised dense vectors $v_q, v_k \in \mathbb{R}^{768}$. As the vectors are unit-normalised, their dot product equals cosine similarity, enabling efficient matrix multiplication scoring. It captures semantic relationships that are invisible to lexical methods (e.g., a query mentioning "glycaemic control" will correctly align with an abstract discussing "blood glucose management" or "HbA1c reduction", whereas a sparse TF-IDF vector would find no overlap between these vocabularies).

[1] Medical Subject Headings (MeSH), U.S. National Library of Medicine: https://www.nlm.nih.gov/mesh/

The abstract cosine score for paper $p_k$ is:

$$\sigma_k^{abs} = v_q \cdot v_k \tag{5}$$

Abstracts provide the richest textual signal, but they can be absent or truncated in PubMed records for older papers, letters or conference abstracts. To maintain scoring quality when abstracts are sparse, a complementary title score based on Jaccard similarity over content token sets $T_q, T_k$ is computed. Stop words and common medical determiners are removed from both sets before the calculation:

$$\sigma_k^{\text{title}} = \frac{|T_q \cap T_k|}{|T_q \cup T_k|} \tag{6}$$

The two components are blended into a *paper semantic relevance score* $\tilde{\sigma}_k$, with abstract cosine receiving the majority weight because it captures semantic depth more reliably than title-level matching.

$$\tilde{\sigma}_k = \left(0.75\, \sigma_k^{\text{abs}} + 0.25\, \sigma_k^{\text{title}}\right) \times \rho_k \tag{7}$$

Where $\rho_k$ represents a recency multiplier applied to prefer recent evidence in fast-moving clinical domains:

$$\rho_k = \begin{cases} 1.15 & \text{if } Y_{\text{now}} - \text{year}(p_k) \le 5 \\ 1.0 & \text{otherwise} \end{cases} \tag{8}$$

The 15% recency boost is intentionally modest to avoid overriding a highly relevant older paper (e.g., a landmark Randomized Controlled Trial-RCT) in favor of a more recent but less informative one. Two guard conditions are enforced at this stage: *i)* papers whose abstract text is fewer than 150 characters (typically title-only stubs or retracted records) are capped at $\tilde{\sigma}_k \le 0.08$ regardless of title similarity, as they provide insufficient textual evidence for claim extraction; *ii)* a hard threshold gate removes all papers with $\tilde{\sigma}_k < \tau_{\text{rel}}$ from the candidate pool. To prevent hypothesis starvation in cases where the literature is genuinely sparse, a minimum number of papers are always retained even if their scores fall below $\tau_{\text{rel}}$.

*3.3.3 Evidence support score*

Beyond individual paper relevance, the overall density of the retrieved literature provides an informative prior on how well-studied a hypothesis is. A hypothesis for which many recent, relevant papers exist is more likely to yield high-quality claims than one for which only a handful of tangentially related records can be found. We operationalize this intuition as the evidence support score:

$$E(H_i) = 1 - \frac{1}{1 + \frac{|Np_i| + 1.5|Np_i^5|}{10}} \tag{9}$$

where $Np_i$-the set of papers retained for hypothesis $H_i$ after the relevance-filtering stage and $Np_i^5 \subseteq Np_i$-the subset published within the last 5 years. The weighting factor 1.5 assigned to $|\, Np_i^5 \,|$ reflects the additional evidential value of recent literature. The denominator scaling constant 10 sets the inflection point so that $E(H_i) \approx 0.5$ when the adjusted paper count reaches 10, being a reasonable threshold for moderate literature support. $E(H_i) \in (0,1)$ follows a sigmoid-like saturation curve that grows quickly for sparse evidence and plateaus for very dense corpora, preventing a single hypothesis from receiving disproportionate evidence inflation simply because its keywords match many records.

*3.3.4 LLM abstract screening gate*

The semantic relevance scoring operates on embedding similarity and does not capture clinical appropriateness. A paper may rank highly because its abstract is topically close to the case description while reporting findings in a completely different clinical context (e.g., a study on the same drug compound but for a different indication, age group or route of administration). This step addresses this with a fast, token-efficient LLM screening call that makes a binary clinical relevance judgment for each (paper, hypothesis) pair before committing to full claim extraction.

The screening prompt provides the LLM with the case summary, the hypothesis text, the abstract capturing the study design statement, the patient population and the primary outcome of paper $p_k$. The LLM returns a binary judgment $r_{i,k} \in \{0,1\}$, where $r_{i,k} = 1$ means the paper is sufficiently relevant to justify claim extraction for hypothesis $H_i$ and $r_{i,k} = 0$ means it should be excluded.

$$f_{\mathcal{L}}(C, H_i, \text{abstract}_k, P_k, O_k) \rightarrow r_{i,k} \in \{0,1\} \quad (10)$$

A fail-open policy is applied: on any LLM response parsing failure (malformed JSON, timeout or refusal), the paper is assigned $r_{i,k} = 1$ and proceeds to claim extraction, ensuring that network instability or ambiguous LLM outputs never silently remove potentially relevant papers. The screening gate is designed to eliminate *obvious* irrelevancies (e.g., wrong population, wrong intervention class, purely methodological papers) rather than making fine-grained relevance distinctions. In practice, this gate typically removes 15–30% of papers that survived the embedding relevance filter, reducing the number of LLM claim-extraction calls and lowering both latency and API cost.

### 3.4 Claim extraction and evidence-level weighting

Claim extraction is the most information-dense stage. Rather than treating each abstract as an atomic relevance signal, SCEPTER decomposes it into a set of atomic, verifiable propositions-*claims*, each representing a single directional relationship supported by the paper's findings. It enables the downstream stages to detect contradictions, compute Pareto fronts and trace every recommendation to specific textual evidence, none of which would be possible if papers were treated as opaque relevance scores.

For each paper $p_k$ that passes the screening gate, the abstract is sent to the LLM together with the structured case $C$, the hypothesis $H_i$ and a structured extraction prompt. The extraction process can be represented as:

$$f_L(C, H_i, \text{abstract}_k) \rightarrow \mathcal{E}_{i,k} = \{e_1, e_2, \dots, e_n\} \quad (11)$$

where $\mathcal{E}_{i,k}$ denotes the set of claims extracted from paper $p_k$ with respect to hypothesis $H_i$. Each claim $e \in \mathcal{E}_{i,k}$ is represented as:

$$e = \langle R, V, rtype, etype, text \rangle \quad (12)$$

where $R \in [0,1]$-*a case-specific relevance score* assigned by the LLM, measuring the extent to which the reported finding applies to the clinical characteristics of the presented patient; $V \in [0,1]$-the LLM-derived methodological *validation score*; $rtype$-the semantic relationship type, $etype$-the study design classification; $text$-the claim statement extracted from the abstract.

The LLM is instructed to extract only claims that are *explicitly* supported by the abstract text, not claims it could infer from background knowledge. The grounding constraint is enforced by including the abstract verbatim in the prompt and asking the LLM to cite the relevant passage. Claims with low thresholds ($R < 0.2$ or $V < 0.1$) are discarded as too weak to influence recommendations, preventing the accumulation of low-quality noise in the evidence pool. Each retained claim carries the PMID, title and publication year of its source paper as provenance metadata.

#### *3.4.1 Evidence-Level Weighting (ELW)*

Validation score $V$ captures intra-paper quality or how well the abstract reports its findings but does not reflect the global evidence hierarchy. A well-written case report can receive a high $V$ score because it describes its methodology and results, while a poorly reported RCT might score lower despite being methodologically superior. To correct it, an evidence-level weight $\omega_e$ is applied as a multiplicative adjustment that anchors each claim's validation score to its study design's position in the Oxford Centre for Evidence-Based Medicine (OCEBM) hierarchy. Thus, the validation score is updated as follows:

$$V \leftarrow \min(\mathrm{V} \times \omega_e,\ 1.0) \quad (13)$$

The ceiling at 1.0 prevents the product from exceeding the valid score range. In practice, $\omega_e = 1.0$ for meta-analyses means that their LLM-reported quality is taken at face value, while $\omega_e = 0.25$ for expert opinion claims means their $V$ score can never exceed one quarter of what the LLM reports, regardless of how confidently the opinion is expressed. It ensures that recommendations based on expert opinion are always presented with lower confidence than those derived from RCTs or systematic reviews, even when the abstract prose is equally authoritative sounding. The ELW in Table 1 follows the OCEBM hierarchy.

Table 1. ELW weights

| Study Design | $\omega_e$ |
|---|---|
| Meta-analysis | 1.00 |
| Systematic review | 0.95 |
| RCT | 0.90 |
| Cohort/Observational | 0.70 |
| Case-control | 0.60 |
| Cross-sectional | 0.55 |
| Case series | 0.40 |
| Case report | 0.35 |
| Expert opinion | 0.25 |
| Unclassified | 0.60 |

*3.4.2 Composite score*

For downstream ranking, deduplication prioritization and Pareto dominance checking, a composite score aggregates the three primary dimensions of a claim's value into a single scalar:

$$\phi(e) = w_R \cdot R + w_E \cdot E(H_i) + w_V \cdot V \quad (14)$$

with default weights $w_R = 0.4$, $w_V = 0.4$, $w_E = 0.2$. The equal weighting of $R$ and $V$ reflects the dual requirement that a useful claim must be both about this specific case and methodologically credible. Neither dimension alone is sufficient. A highly relevant but methodologically weak claim (e.g., a case report perfectly matching the patient's profile) should not outrank a high-quality systematic review that is moderately less specific to this case. The 20% evidence score contribution ensures that claims from hypotheses that are well-supported by literature receive a modest boost, aligning individual claim scoring with the hypothesis-level reasoning. $\phi(e)$ is used as an *ordering* and *tiebreaking* metric rather than as a final quality score. The three score components originate from complementary sources: $R$ captures case-specific relevance estimated by the LLM, $V$ captures methodological credibility adjusted by the evidence hierarchy and $E(H_i)$ reflects the degree of support for the hypothesis in the retrieved literature corpus.

**3.5 Claim deduplication and contradiction detection**

While PubMedBERT dense embeddings are used to recover semantically related abstracts across vocabulary mismatches, TF-IDF is used for claim-level deduplication and contradiction detection. Claims are short, structurally uniform outputs produced by the same LLM under the same prompt schema, which means their vocabulary is largely controlled and consistent, meeting the conditions under which sparse lexical similarity performs well. Dense embeddings provide marginal gains at the cost of additional inference calls on a typically small set (fewer than 50 claims per run). More importantly, the fine-grained semantic distinctions that dense representations capture (e.g., "reduces mortality" versus "improves survival") are already handled by the explicit LLM-based relationship scoring.

*3.5.1 Claim deduplication*

When multiple PICO queries for the same hypothesis retrieve overlapping papers or when different papers report the same fundamental finding in similar language, near-identical claims accumulate in the evidence pool, being common for well-established clinical relationships that appear in review papers, textbooks and primary studies simultaneously. Without deduplication, these near-identical claims inflate the apparent strength of evidence, causing the Pareto front and consensus scores to be dominated by the most frequently cited finding rather than the most informative one. Deduplication ensures that each distinct evidential proposition is counted once, regardless of how many papers report it.

For this task, a TF-IDF vectorizer is fitted over all extracted claim texts $\{text(e_k)\}$ and pairwise cosine similarities are computed. Claims $e_a, e_b$ are considered duplicates when:

$$\cos(u_a, u_b) \geq \tau_{dup} \quad (15)$$

where $u_a, u_b$-the TF-IDF vector representations of claims $e_a$ and $e_b$, respectively, $\cos(u_a, u_b)$-cosine similarity.

The threshold $\tau_{\mathrm{dup}} = 0.8$ is set empirically to distinguish near-verbatim repetitions from claims that address the same topic with different nuances. Before the greedy pass, all claims are sorted by $\phi(\cdot)$ in descending order so that the canonical representative of each duplicate group is always the highest-scoring one. For each canonical claim, all duplicate sources are merged into its provenance set, so the final claim still carries multi-paper citation support even after deduplication. A claim supported by five different papers after deduplication is treated as stronger evidence than one supported by a single paper, even though both are represented by a single claim object in the pool. Deduplication is applied once globally across all hypotheses and all claims before Pareto analysis. Applying it globally rather than per-hypothesis is important because the same finding may arise under different hypotheses.

*3.5.2 Contradiction detection*

Conflicting evidence is clinically important information since it tells the practitioner that a relationship is contested in the literature and that acting on a single study would be risky. Rather than hiding contradictions by averaging or overriding them, SCEPTER surfaces them explicitly as a first-class output. The challenge is distinguishing evidential contradictions from unrelated claims that happen to use opposing verbs or from claims that are simply discussing different aspects of the same topic at different levels of specificity.

SCEPTER uses a two-condition gate that requires both topical similarity *and* relationship polarity mismatch. Two non-duplicate claims $e_a, e_b$ are flagged as contradictory when:

$$\tau_{topic} \leq \cos(u_a, u_b) < \tau_{dup} \wedge rtype(e_b) \in \mathrm{Opp}(rtype(e_a)) \quad (16)$$

where $\tau_{topic}$-the duplicate topic detection threshold set to 0.5; $rtype$ $(\cdot)$-the claim relationship type; *Opp(·)*-the set of relationship types that are directionally incompatible with it. Thus, positive-effect relationships (e.g., *increases*) are considered contradictory to negative-effect (*decreases*) and null-effect (*no_effect*) relationships, while *decreases* similarly contradict *increases* and *no_effect*. For associative relationships such as *associated*, *predicts*, *moderates* and *diagnostic*, only a reported *no_effect* finding is treated as contradictory. Also, *no_effect* is considered incompatible with any relationship indicating a measurable association or directional effect. The cosine similarity window $[\tau_{topic} = 0.5, \tau_{dup} = 0.8]$ defines the regime of *same-topic, non-identical* claims. Claims below 0.5 are topically too distant to constitute a meaningful contradiction (they may simply discuss different aspects of the case), while claims above 0.8 are de facto duplicates and should not be treated as conflicts even if their relationship labels differ.

Contradictions supported by higher-quality study designs are therefore ranked as more significant than conflicts arising from weaker forms of evidence. The resulting contradiction set is incorporated into the evidence synthesis stage, where conflicting findings are explicitly reported and contextualized for the clinician.

### 3.6 Pareto-optimal claim selection

After deduplication, the framework identifies which claims should be elevated as primary evidence for recommendations. A naive approach would sort by $\phi(e)$ and select the top $N$ claims. However, this discards the inherent tension between the three scoring dimensions. A claim with $R = 0.95$, $V = 0.40$, $E = 0.60$ might score identically to one with $R = 0.60$, $V = 0.80$, $E = 0.55$ under a weighted average, yet they represent fundamentally different trade-offs. The first claim is highly case-specific but comes from weaker evidence, the second comes from stronger methodological evidence but is moderately less specific. Neither is strictly better and both should inform the clinician's decision.

We address this by framing claim selection as a three-objective maximization problem over the objective space $\Omega = [0,1]^3$, where each claim $e$ is represented as a point $f(e) = (R_e, E_e, V_e) \in \Omega$.

Claim $e_a$ strictly dominates claim $e_b$ (denoted by $e_a \succ e_b$) if and only if $e_a$ is at least as good on every objective and strictly better on at least one:

$$e_a \succ e_b \iff R_a \geq R_b \wedge E_a \geq E_b \wedge V_a \geq V_b \wedge f(e_a) \neq f(e_b) \quad (17)$$

The dominance count of a claim $e$ is the number of claims that dominate it:

$$d(e) = |\{e' \in \mathcal{E} : e' \succ e\}| \quad (18)$$

The Pareto-optimal set $\mathcal{P} \subseteq \mathcal{E}$ is the set of claims with zero dominance count or the non-dominated frontier:

$$\mathcal{P} = \{e \in \mathcal{E} \mid d(e) = 0\} \quad (19)$$

Claims not in $\mathcal{P}$ are Pareto-dominated. There exists at least one other claim that is superior on every evaluated dimension. To provide a complete ranking of all claims beyond the binary Pareto-optimal flag, a full layered Pareto ranking is available. Pareto rank 1 is the front $\mathcal{P}$. Rank 2 is the front computed after removing $\mathcal{P}$ from $\mathcal{E}$ and so on iteratively:

$$\mathcal{P}^{(1)} = \mathcal{P}, \mathcal{P}^{(k)} = \mathrm{ParetoFront}\left(\mathcal{E} \setminus \bigcup_{j=1}^{k-1} \mathcal{P}^{(j)}\right) \quad (20)$$

This stratification is analogous to the non-dominated sorting used in NSGA-II and produces a total ordering of the evidence pool from most to least multi-dimensionally superior. The recommendation stage uses Rank 1 claims as primary evidence. Rank 2 claims serve as supporting references when Rank 1 is insufficient.

### 3.7 Consensus analysis

The Pareto front identifies the best individual claims, but it does not convey the overall direction of the evidence for each hypothesis. A clinician needs to know not only which claims are strongest, but whether most of the evidence supports or contradicts a given hypothesis and how strongly. This stage computes a per-hypothesis consensus statistics object that captures this aggregate signal.

For each hypothesis $H_i$, all non-duplicate claims assigned to that hypothesis are classified by their relationship type into three groups:

- Supportive ($n_{\text{sup}}$)-claims with relationships in *{increases, associated_with, predicts, moderates, diagnostic}* when the direction aligns with the hypothesis;
- Opposing ($n_{\text{opp}}$)-claims with relationships in {*decreases, no_effect*} or any relationship that contradicts the hypothesis direction;
- Neutral ( $n_{\text{neutral}}$ )-claims whose relationship type cannot be unambiguously classified as supportive or opposing in the hypothesis context.

A directional consensus score is computed as the proportion of classified claims that agree on a single direction:

$$\gamma(H_i) = \frac{\max(n_{\text{sup}}, n_{\text{opp}})}{n_{\text{sup}}+n_{\text{opp}}+n_{\text{neutral}}} \tag{21}$$

where $\gamma(H_i) \in (0,1]$ with $\gamma = 1$ indicates perfect unanimity and $\gamma \approx 0.5$ indicates an equal split between supportive and opposing claims.

The direction label is set to *supportive* if $n_{\text{sup}} > n_{\text{opp}}$, *opposing* if $n_{\text{opp}} > n_{\text{sup}}$ and *mixed* when the counts are equal. Hypotheses are ranked by $\gamma(H_i)$ in the output, providing the clinician with a clear ordering from best-supported to most contested. This ranking, along with the raw counts and direction label, is passed to the evidence synthesis stage where it structures the narrative and to the recommendation stage where it informs the assigned evidence level of each recommendation.

## 3.8 Recommendation and paper Q&A

### *3.8.1 Evidence synthesis and recommendations*

The structured outputs produced by SCEPTER are individually useful but collectively dense. A clinician working under time pressure needs a concise, readable narrative that integrates these signals into a coherent clinical picture. This step generates this narrative using a structured LLM synthesis prompt. The LLM is provided with the following: *i)* the top Pareto-optimal claims serialized with their $R, E, V$ scores; *ii)* all detected contradiction groups with their source PMIDs and study design labels; *iii)* per-hypothesis consensus direction and $\gamma$ scores. The LLM is instructed to produce a structured four-section narrative:

1. *Consistent findings*-Clinical relationships for which multiple independent studies, ideally of different designs, converge on the same directional conclusion. It anchors the recommendations in replicable evidence.
2. *Contested evidence*-Specific contradictions detected, explained with reference to the methodological differences (study design, population, intervention dosing) that might account for the divergence. Rather than dismissing contradictions, it contextualizes them.
3. *Evidence gaps*-Relationships implied by the hypothesis set or clinical context that lack direct empirical evidence in the retrieved literature, including missing comparator arms, under-studied sub-populations or outcome measures that are frequently studied in isolation but not jointly;
4. *Overall evidence quality*-A summary of the study design distribution across all retained claims (e.g., proportion of RCTs vs. observational studies) and an explicit statement of the overall confidence level the clinician should place in the synthesis.

The synthesis prompt is designed so that all factual claims in the narrative must be traceable to a specific PMID from the input. The LLM is instructed to append PMID citations inline and to avoid generating unsupported assertions. If the Pareto-optimal claim pool is smaller than expected (e.g., due to very sparse retrieval), the LLM is instructed to explicitly acknowledge the limitation rather than generating confident-sounding but weakly grounded text. A rule-based fallback synthesis is triggered if the LLM call fails, producing a structured plain-text summary directly from the scoring data.

The final step translates the curated evidence pool into actionable clinical recommendations. The LLM is prompted with the Pareto-optimal claims grouped by hypothesis, the evidence synthesis narrative, the consensus direction and confidence levels. It is instructed to generate up to $N_R = 5$ distinct, actionable recommendations, where each recommendation must correspond to a clinical action (diagnostic test, therapeutic intervention, monitoring strategy or referral decision) that is directly supported by at least one Pareto-optimal claim.

Each recommendation is assigned to an evidence level according to the study-design composition of its supporting claims. Recommendations are classified as *Strong* when supported by at least two RCTs or meta-analyses. A *Moderate* evidence level is assigned when the supporting evidence is dominated by cohort, case-control or other observational studies. Recommendations are classified as *Weak* when they rely primarily on case reports, expert opinion or otherwise limited evidence. This classification mirrors the general hierarchy of evidence used in evidence-based medicine and provides clinicians with an interpretable measure of recommendation reliability.

The LLM is explicitly prohibited from generating recommendations that are not supported by claims in the Pareto-optimal set, even if it had domain knowledge that would suggest a plausible action, grounding the system's outputs in the retrieved evidence and preventing the LLM from hallucinating references to studies that were not retrieved or did not pass the quality filters.

The recommendations and synthesis narrative provide an aggregated view of the evidence, but a specialist may wish to interrogate specific papers directly, for example to determine whether a study's inclusion criteria apply to the current patient or to examine reported effects for a clinically relevant sub-group. To support it, SCEPTER includes a *Paper Q&A* module based on RAG. The module allows users to select one or more supporting publications and ask natural-language questions grounded in the retrieved scientific evidence. The selected documents are combined with the structured case representation $C$, enabling the LLM to generate patient-specific answers while remaining anchored to the source literature.

*3.8.2 Evidence provenance*

To ensure transparency and auditability, SCEPTER maintains provenance information throughout the entire pipeline. Every extracted claim retains metadata linking it to its source publication, including the PMID, title, publication year, journal and study design. These metadata are propagated during evidence-level weighting, claim deduplication, contradiction detection, Pareto selection and recommendation generation.

Formally, every recommendation $r_j$ must be traceable to at least one supporting claim $e_i$ and every claim must originate from at least one scientific publication $p_k$:

$$r_j \leftarrow e_i \leftarrow p_k \tag{22}$$

When duplicate claims are merged, the provenance records of all supporting papers are preserved, allowing recommendations to reflect evidence originating from multiple independent sources. Recommendations that cannot be linked to at least one supporting claim and one source publication are excluded from the final output. The same provenance mechanism is used by the Paper Q&A module. Answers generated in response to specialist queries remain grounded in the retrieved literature and include references to the supporting publications, enabling direct verification of the underlying evidence.

## 4. Simulations and results

### 4.1 Implementation details

The pipeline is implemented in Python using *google-genai* for LLM calls (Google Gemini 3 Flash), *biopython* for NCBI Entrez API access, *sentence-transformers* for PubMedBERT dense embeddings, *scikit-learn* for TF-IDF vectorization. All extraction and scoring prompts are executed with temperature set to zero to maximize output determinism and reduce variability across repeated runs. Also, LLM calls use a low temperature for narrative generation to maximize output determinism. Every LLM call specifies a structured JSON output schema and includes a regex-based extraction fallback that attempts to parse JSON embedded within prose responses. This two-layer parsing approach handles the frequent case where a model produces valid JSON wrapped in markdown code fences or preceded by a brief explanation sentence. The hyperparameters are set as in Table 2.

Table 2. Summary of key hyperparameters

| Parameter | Symbol | Default | Description |
|---|---|---|---|
| Hypotheses per case | $N_H$ | 3 | LLM-generated hypotheses |
| PICO queries per hypothesis | $N_P$ | 3 | PubMed query variants |
| Papers per PICO query | $K_q$ | 30 | Max PubMed IDs fetched |
| Relevance threshold | $\tau_{\text{rel}}$ | 0.20 | Min $\tilde{\sigma}_k$ to keep paper |
| Top papers kept per hypothesis | $K$ | 20 | After relevance filter |
| Deduplication threshold | $\tau_{\text{dup}}$ | 0.8 | Cosine threshold for duplicate |

| Contradiction lower bound | $\tau_{\text{topic}}$ | 0.5 | Cosine threshold for same topic |
|---|---|---|---|
| Composite weights | $w_R, w_V, w_E$ | 0.4, 0.4, 0.2 | Claim scoring weights |
| Min claim relevance | $R_{\min}$ | 0.20 | Discard threshold for $R$ |
| Min claim validation | $V_{\min}$ | 0.10 | Discard threshold for $V$ |
| Max recommendations | $N_R$ | 5 | Per case |
| Full-text PMC papers (Q&A) | - | 3 | Max open-access papers fetched |
| Max chars per paper (Q&A) | - | 48 000 | Context cap per PMC document |

## 4.2. Evaluation metrics

### *4.2.1 Compression ratios*

To quantify the efficiency of SCEPTER, we define a set of compression ratios that measure how efficiently the framework transforms the initial PubMed search space into more compact evidence representations. The compression is evaluated at three levels: retained papers after relevance filtering; Pareto-optimal claims after multi-objective selection; and final clinical recommendations. The retrieval compression ratio measures the reduction from the initial PubMed search results to the paper set retained after relevance filtering:

$$CR_{\text{PubMed}\to\text{retrieved}} = \frac{N_{\text{PubMed}}}{N_{\text{retrieved}}} \tag{23}$$

where $N_{\text{PubMed}}$-the number of papers returned by the initial PubMed search, $N_{\text{retrieved}}$-the number of papers retained after semantic ranking and relevance filtering.

The Pareto compression ratio measures the reduction from the initial PubMed search results to the final Pareto-optimal claim set:

$$CR_{\text{PubMed}\to\text{Pareto}} = \frac{N_{\text{PubMed}}}{N_{\text{Pareto}}} \tag{24}$$

where $N_{\text{Pareto}}$-the number of non-dominated claims belonging to the Pareto front.

The recommendation compression ratio measures the reduction from the initial PubMed search results to the final recommendation set:

$$CR_{\text{PubMed}\to\text{recommendations}} = \frac{N_{\text{PubMed}}}{N_{\text{recommendations}}} \tag{25}$$

where $N_{\text{recommendations}}$-the number of recommendations generated by the framework.

Higher compression-ratio values indicate a greater reduction of the original literature space while preserving the evidence required for recommendation generation.

### *4.2.2 Pareto hypervolume*

The size of the Pareto front alone does not convey how well the evidence pool covers the objective space. A front consisting of 10 claims clustered in a small region is less informative than a front of 5 claims spread across the full $[0,1]^3$ cube. The hypervolume indicator $HV$ measures the volume of the objective space dominated by the Pareto front:

$$HV(\mathcal{P}) = \lambda_3\left(\bigcup_{e\in\mathcal{P}}[R_e, 1] \times [V_e, 1] \times [E_e, 1]\right) \tag{26}$$

where $\lambda_3$-the three-dimensional Lebesgue measure. $HV \in [0,1]$ with $HV = 1$ indicates better covering the entire unit cube and $HV \to 0$ indicates a small front concentrated near the origin. In practice, $HV$ is reported as a pipeline metadata metric that helps a researcher assess the overall quality of the retrieved evidence for a given case. A high hypervolume indicates that the pipeline found claims covering diverse quality profiles, while a low hypervolume suggests that the evidence is uniformly weak in one or more dimensions.

### *4.2.3 Evidence Type Entropy* (ETE)

Entropy measures the diversity of study designs across extracted claims. The standard definition is Shannon entropy over the empirical distribution of evidence types:

$$ETE = -\sum_{t\in etype} e_t \log_2 p_t \tag{27}$$

where $etype = \{\text{RCT, cohort, meta-analysis, case report, ...}\}$ is the set of observed evidence types and $e_t = \frac{n_t}{\sum_{t'} n_{t'}}$ is the proportion of claims with evidence type $t$. $ETE = 0$ means all claims come from a single study design. Higher values indicate a more balanced mix of study designs in the retained evidence pool.

## 4.3 Aggregate performance across clinical cases

To evaluate SCEPTER, we conducted experiments on a set of 150 clinical cases collected from a university psychological counselling context. The cases focused on common student mental-health challenges, including anxiety, stress, sleep disturbances, academic burnout, concentration difficulties and related psychosocial issues arising in an academic environment. The selected cases represent realistic scenarios frequently encountered by university psychologists, where multiple interacting factors may contribute to the patient's presentation and where the relevant scientific literature spans diverse psychological, behavioral and clinical domains. Table 3 summarizes the aggregate performance metrics averaged across all 150 cases, including retrieval efficiency, claim extraction statistics, evidence-quality indicators, Pareto-front characteristics and recommendation generation outcomes.

Table 3. Performance metrics across 150 case studies

| Metric | Mean over 150 cases |
|---|---|
| Papers found in the last 10 years | 576 |
| Papers retrieved | 53 |
| Claims extracted | 69 |
| Duplicate claims | 4 |
| Contradiction pairs | 6 |
| Pareto-optimal claims | 7 |
| Recommendations | 3 |
| *R* score | 0.785 |
| *V* score | 0.517 |
| *E* score | 0.824 |
| Composite φ score | 0.721 |
| *ETE* | 0.901 |
| Pareto *HV* | 0.21 |
| H1 claims / yield | 20 / 1.43 |
| H2 claims / yield | 26 / 1.30 |
| H3 claims / yield | 19 / 1.58 |

In terms of evidence reduction, the average PubMed search returned 576 papers published within the last 10 years. After semantic ranking and relevance filtering, it was reduced to 53 retained papers, corresponding to a retrieval compression ratio $CR_{\text{PubMed}\rightarrow\text{retrieved}} = 10.9:1$. The subsequent claim extraction, deduplication and Pareto-selection stages further reduced the evidence space to an average of 7 Pareto-optimal claims, yielding a Pareto compression ratio $CR_{\text{PubMed}\rightarrow\text{Pareto}} = 82.3:1$. The framework generated an average of 3 recommendations per case, corresponding to an overall paper-to-recommendation compression ratio $CR_{\text{PubMed}\rightarrow\text{recommendations}} = 192:1$.

Despite this reduction, the retained evidence maintained high diversity ($ETE = 0.901$) and a non-trivial Pareto hypervolume ($HV = 0.21$), indicating that the final recommendations are derived from claims spanning different combinations of relevance, methodological quality and literature support rather than from a narrowly concentrated evidence pool.

The high mean relevance score ($R = 0.785$) indicates that the extracted claims are generally well aligned with the case context, while the moderate validation score ($V = 0.517$) reflects the heterogeneous methodological quality typical of clinical literature. The high evidence support score ($E = 0.824$) suggests that most hypotheses are supported by dense recent literature.

### 4.4 Experts evaluation

To assess the practical utility of SCEPTER, an expert evaluation study was conducted involving five psychologists affiliated with a university psychological counselling and research center. Each expert independently reviewed up to 150 case descriptions and evaluated the generated recommendations using a structured questionnaire based on Likert-scale criteria (Table 4). The evaluation focused on clinical relevance, evidence adequacy, explainability, actionability, trustworthiness, contradiction usefulness, time-saving potential and utility.

Table 4. Expert evaluation criteria for SCEPTER recommendations

| Criterion | Evaluation Question | Scale | Score |
|---|---|---|---|
| Clinical Relevance | *How relevant are the recommendations to the presented clinical case?* | 1 (Not relevant)-5 (Highly relevant) | 4.5 |
| Evidence Adequacy | *Does the supporting evidence adequately justify the recommendations?* | 1 (Very inadequate)-5 (Very adequate) | 4.9 |
| Explainability | *Is the rationale behind the recommendations clearly explained and easy to understand?* | 1 (Very unclear)-5 (Very clear) | 4.8 |

| Actionability | *Can the recommendations be translated into concrete clinical actions?* | 1 (Not actionable)-5 (Highly actionable) | 4.5 |
|---|---|---|---|
| Trustworthiness | *How much would you trust these recommendations when making a clinical decision?* | 1 (No trust)-5 (High trust) | 4.8 |
| Contradiction Usefulness | *Was the presentation of contradictory findings useful for clinical interpretation?* | 1 (Not useful)-5 (Very useful) | 4.9 |
| Time-Saving Potential | *Compared with manually reviewing the literature, how much time would this output save?* | 1 (No saving)-5 (Very significant saving) | 5.0 |
| Overall Utility | *How useful would this system be in your routine clinical workflow?* | 1 (Not useful)-5 (Extremely useful) | 4.9 |

Table 4 indicates a positive expert assessment across all evaluation dimensions. The highest score was obtained for *Time-Saving Potential* (5.0/5), suggesting that the experts perceived practical value in the framework's ability to reduce the effort associated with manual literature review. *Contradiction Usefulness* (4.9/5), *Evidence Adequacy* (4.9/5) and *Overall Utility* (4.9/5) also received high ratings, indicating that the experts valued both the quality of the supporting evidence and the framework's contribution to clinical decision support. *Clinical Relevance* (4.5/5) and *Actionability* (4.5/5) received slightly lower, although still highly positive, evaluations, reflecting the inherent variability of individual clinical cases and the challenge of translating literature-derived evidence into universally applicable interventions

**4.5 Representative clinical case and parameter sensitivity analysis**

To characterize the sensitivity of the pipeline to the relevance threshold $\tau_{\mathrm{rel}}$ and to validate the Pareto front behavior under realistic retrieval conditions, we executed three controlled runs on a single insomnia-comorbid-anxiety clinical case that describes a 22-year-old university student presenting with a three-month history of progressive anxiety that significantly worsened during the examination period. Clinical features include: *i) psychological*-excessive worrying about academic performance, fear of failure, constant mental fatigue, feelings of inadequacy relative to peers and moderate depressive symptoms on psychometric assessment, no suicidal ideation; *ii) behavioral*-reduced social interactions, partial class avoidance, procrastination, low academic productivity, cessation of physical activity and irregular eating habits; *iii) somatic*-palpitations, muscle tension, recurrent headaches, gastrointestinal discomfort, triggered by deadlines or presentations; *iv) sleep*-severe disturbance characterized by difficulty initiating sleep (onset delayed to 02:00–03:00) and frequent nocturnal awakenings; *v) substance use*-caffeine intake escalated from 1 to 5 cups per day over the preceding six weeks; *vi) history*-one prior episode of short-term counselling during the first academic year, no current psychiatric medication.

Psychometric assessments indicated elevated generalized anxiety, high perceived stress and severe sleep disturbance. The clinician's initial differential diagnosis is Generalized Anxiety Disorder (GAD) or adjustment disorder with mixed anxiety and depressed mood, with academic burnout and high caffeine consumption identified as probable exacerbating factors. The case is chosen for simulation because it presents a realistic multi-factorial presentation where three meaningfully distinct hypotheses can be formulated, each pulling toward a different body of literature. The three hypotheses generated by the pipeline are:

- H1 (mechanistic)-Caffeine consumption of approximately 400–500 mg/day exacerbates somatic anxiety symptoms (palpitations, muscle tension) and prolongs sleep onset latency through adenosine receptor antagonism and elevated sympathetic nervous system activity.
- H2 (intervention)-A six-week CBT for insomnia (CBT-I) intervention combined with standard CBT for anxiety produces a greater reduction in generalized anxiety and depressive symptoms compared to standard CBT for anxiety alone.
- H3 (diagnostic)-Standardized evaluation for adult ADHD reveals a primary inattentive ADHD phenotype rather than primary GAD, given the chronic procrastination, concentration difficulties and academic anxiety pattern.

All runs used identical parameters (Table 2) except $\tau_{\mathrm{rel}}$, which is varied across $\{0.15, 0.20, 0.30\}$. Table 5 compares the results of the $\tau_{\mathrm{rel}}$ scenarios.

Table 5. Cross-threshold simulation results

| Metric | $\tau_{\text{rel}} = 0.15$ | $\tau_{\text{rel}} = 0.20$ | $\tau_{\text{rel}} = 0.30$ |
|---|---:|---:|---:|
| Elapsed time (s) | 557.5 | 389.4 | 328.6 |
| Papers retrieved | 78 | 60 | 46 |
| Claims extracted | 85 | 67 | 65 |
| Duplicate claims | 2 | 1 | 0 |
| Contradiction pairs | 1 | 2 | 8 |
| Pareto-optimal claims | 9 | 6 | 2 |
| Recommendations | 3 | 3 | 2 |
| *R* mean / std | 0.681 / 0.135 | 0.771 / 0.119 | 0.754 / 0.132 |
| *V* mean / std | 0.449 / 0.176 | 0.464 / 0.158 | 0.492 / 0.169 |
| *E* mean / std | 0.853 / 0.058 | 0.925 / 0.024 | 0.772 / 0.090 |
| Composite φ mean / std | 0.623 / 0.097 | 0.679 / 0.081 | 0.653 / 0.095 |
| *ETE* | 0.841 | 1.391 | 0.888 |
| Pareto *HV* | 0.316 | 0.153 | 0.067 |

Increasing $\tau_{\text{rel}}$ from 0.15 to 0.20 raised the mean composite score from $\phi = 0.623$ to $\phi = 0.679$ and the mean $R$ from 0.681 to 0.771, confirming that stricter semantic pre-filtering does improve per-claim relevance. The evidence type entropy also increased from 0.841 to 1.391 at $\tau_{\text{rel}} = 0.20$, reflecting a more balanced mix of study designs (RCT, observational, systematic review, expert opinion, case report all present), suggesting that at the lower threshold, lower-quality observational papers dominate by sheer volume.

Despite higher per-claim quality metrics, the run at $\tau_{\text{rel}} = 0.30$ produced the worst Pareto outcomes: only 2 Pareto claims, a hypervolume of 0.067 (versus 0.316 at $\tau_{\text{rel}} = 0.15$), being attributable to *objective-space clustering*. The tighter semantic filter selects papers from a narrow topical region, producing claims with homogeneous $(R, V, E)$ coordinates. In a dense, homogeneous cluster, nearly every claim is dominated by at least one neighbour, leaving a very small non-dominated front. Thus, the hypervolume collapse is a direct consequence of reduced evidence diversity rather than reduced evidence quality.

The contradiction count grew from 1 at $\tau_{\text{rel}} = 0.15$ to 8 at $\tau_{\text{rel}} = 0.30$. Counterintuitively, the higher contradiction count at $\tau_{\text{rel}} = 0.30$ reflects greater topical coherence rather than noisier retrieval. Semantically similar papers discussing the same clinical relationship from different study designs are more likely to produce claims with opposing relationship labels. At $\tau_{\text{rel}} = 0.15$, off-topic papers dilute the claim pool, reducing the chance that any two claims are simultaneously topically close and directionally opposed.

The Sankey diagram in Figure 2 illustrates how evidence is progressively condensed throughout the SCEPTER pipeline. Starting from 3 hypotheses, 60 retrieved papers produced 67 claims, of which 6 remained on the Pareto front after multi-objective selection. These Pareto-optimal claims form the evidential basis for the 3 final recommendations. The diagram highlights the contribution of each hypothesis to the retained evidence and demonstrates the traceability of recommendations to specific claim groups.

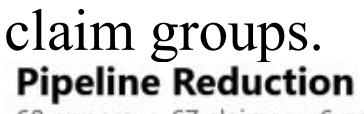

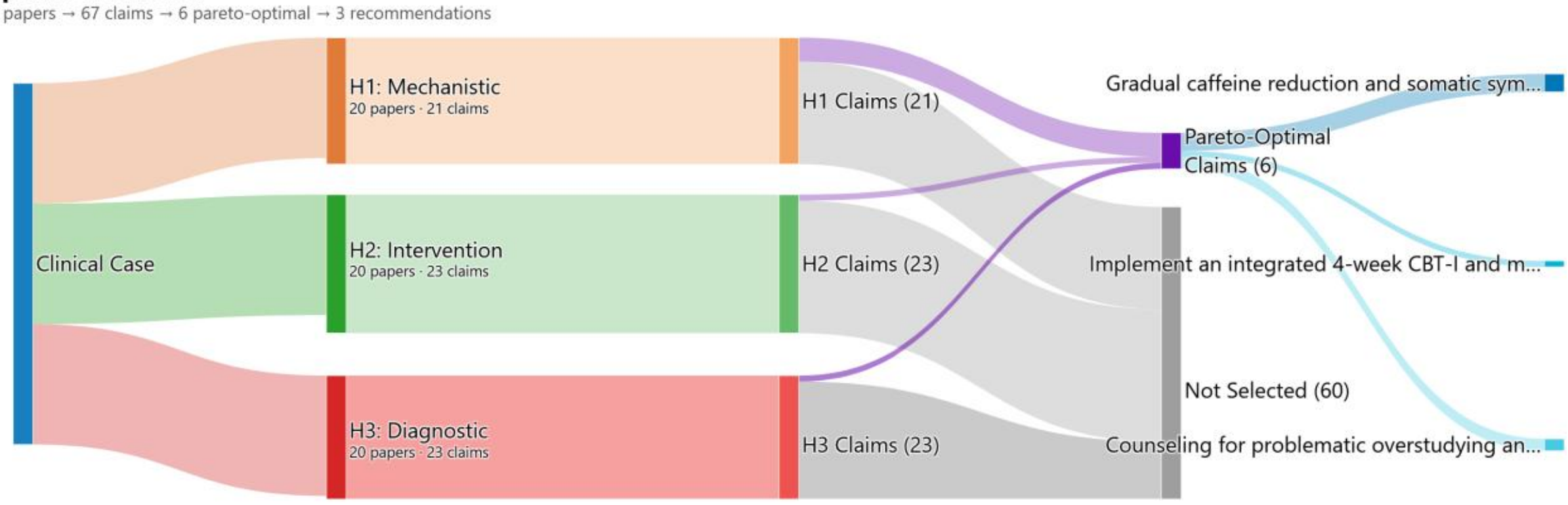

Figure 2. Reduction flow from hypotheses to recommendations for the representative clinical case ($\tau_{rel} = 0.20$)

Figure 3 illustrates the distribution of extracted claims in the $(R, V)$ objective space for the representative clinical case. As can be noticed, no single claim simultaneously maximizes relevance, validation and evidence support. Instead, the Pareto front preserves multiple high-quality claims representing different trade-offs between these objectives. The distribution confirms that the Pareto selection mechanism retains evidence diversity across competing hypotheses rather than favoring a single high-scoring claim, thereby providing a broader and more balanced evidential basis for recommendation generation.

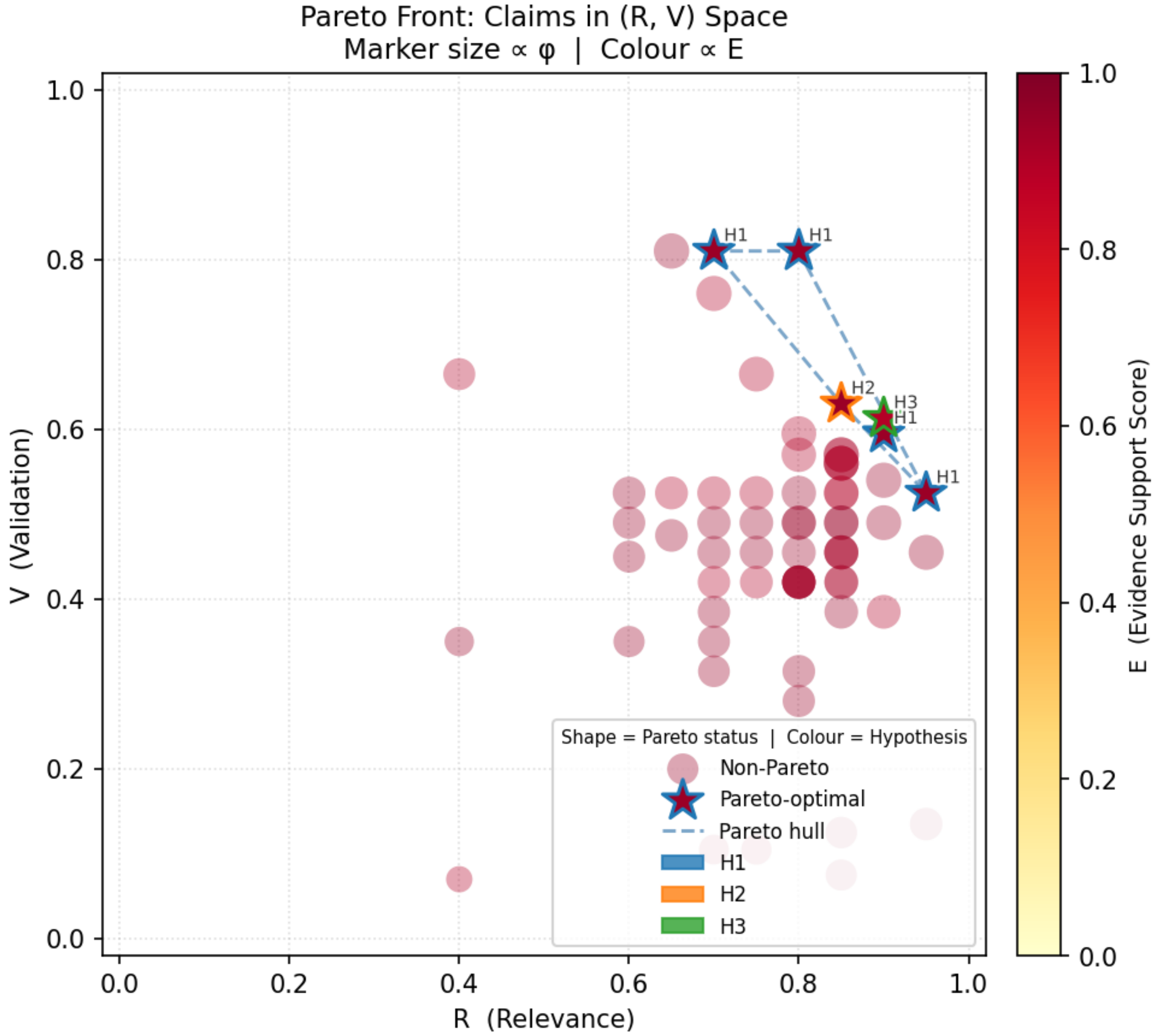

Figure 3. Pareto front with the optimal claims of the case study ($\tau_{\text{rel}} = 0.20$)

To illustrate the utility of the Paper Q&A module, the specialist selected 3 publications associated with the caffeine–sleep–anxiety hypothesis (Figure 4) and asked: *"To what extent does the literature support reducing caffeine consumption as a clinically relevant intervention for this student's anxiety and sleep disturbances, and what level of evidence supports this recommendation?"* The module generated a case-specific synthesis and assessed the strength of the available evidence. The response identified consistent associations between increased caffeine consumption, elevated anxiety symptoms, poorer sleep quality and reduced academic performance among university students. It further highlighted dose-dependent effects, showing that moderate coffee consumption (1–2 cups/day) was associated with better outcomes than heavy consumption (≥3 cups/day). The module also identified methodological limitations of the evidence base, noting that the supporting studies presented in the selected papers were primarily observational and therefore insufficient to establish direct causality.

To what extent does the literature support reducing caffeine consumption as a clinically relevant intervention for this student's anxiety and sleep disturbances, and what level of evidence supports this recommendation?

Based on the provided literature, there is strong observational support for reducing this student's caffeine consumption (currently 5 cups of coffee/day) to a moderate level (1–2 cups/day) to address their anxiety, sleep disturbances, and physical symptoms. However, the **level of evidence supporting this recommendation is low (Level IV)**, as the supporting studies are entirely cross-sectional and observational, establishing correlation rather than direct causality.

**Supporting Details**

**1. Association with Anxiety and Sleep Disturbances**

- **Increased Anxiety and Insomnia:** [Paper 2] (PMID 38674822) demonstrated that university students who increased their coffee consumption had significantly higher scores on the State-Trait Anxiety Inventory (STAI) and the Insomnia Severity Index (ISI).
- **Physical Symptom Correlation:** The student's physical complaints (palpitations, headaches, gastrointestinal discomfort) align directly with the documented adverse effects of excessive caffeine noted in [Paper 2], which include tachycardia, headaches, and gastrointestinal disorders.

**2. Dose-Dependent Effects and Academic Performance**

- **Heavy vs. Moderate Consumption:** [Paper 3] (PMID 41841082) found a dose-related association between coffee intake and student outcomes. While moderate consumption (1–2 cups/day) was associated with higher examination scores, heavy consumption ($\geq 3$ cups/day)—matching the student's current intake of 5 cups—was associated with poorer academic performance.
- **Sleep Deprivation and Anxiety Symptoms:** In [Paper 3], coffee-consuming students who experienced significant sleep/anxiety symptoms or sleep deprivation ($< 5$ hours/night) scored significantly lower on objective examinations. The paper notes that the short-term stimulatory benefits of caffeine are often offset by anxiety and sleep disruption when consumed excessively.

**3. Limitations in the Level of Evidence**

- **No Causal Proof:** Both [Paper 2] (PMID 38674822) and [Paper 3] (PMID 41841082) explicitly state that their **cross-sectional designs** prevent them from establishing a causal relationship.
- While the literature strongly suggests that the student's high caffeine intake is exacerbating their clinical presentation, these papers do not provide interventional data (such as a randomized controlled trial) proving that caffeine reduction will directly resolve the symptoms.

*(Note: [Paper 1] [PMID 41823929] focuses on prescription stimulant shortages and notes that 26.5% of students facing shortages use other stimulants like caffeine to cope, but it does not directly address caffeine reduction as a therapeutic intervention).*

Sources used

PMID 41823929 – University Students and the Stimulant Shortage: A Survey of Strategies to Obtain (2026) – *full text (PMCID 13200170)*

PMID 38674822 – Change in Caffeine Consumption after Pandemic (CCAP-Study) among University Stud (2024) – *full text (PMCID 11053568)*

PMID 41841082 – Association Between Coffee Consumption and Academic Performance Among Second-Yea (2026) – *full text (PMCID 12989166)*

Figure 4. Sample Q&A conversation based on 3 selected papers

### 4.6 Ablation study-impact of multi-objective evidence selection

To evaluate the contribution of the main architectural components, we conduct an ablation study comparing three variants: *i) Baseline* performs PubMed retrieval without filtering the papers by the relevance score and generate recommendations using the composite score $\phi$ to rank claims, without deduplication, contradiction analysis, consensus computation or Pareto selection. The retrieval stage collects up to $N_H \times N_Q \times K_q$ PubMed records, corresponding to $3 \times 3 \times 30 = 270$ candidate records per case before PMID deduplication; *ii) Intermediate* extends the *Baseline* by incorporating relevance score filtering, claim deduplication and contradiction detection, but replaces Pareto selection with a simple top-$K$ ranking based on $\phi$; *iii) Full* employs the complete SCEPTER pipeline, including multi-objective Pareto-optimal claim selection. The comparison results are presented in Table 6.

Table 6. Results of the ablation study

| **Metric** | **Baseline** | **Intermediate** | **SCEPTER (Full pipeline)** |
|---|---|---|---|
| Papers retrieved | 270 | 53 | 53 |
| Claims extracted | 283 | 69 | 69 |
| Duplicate claims | 25 | 4 | 4 |
| Contradiction pairs | 36 | 6 | 6 |
| Selected claims | 20 (ranking based on $\phi$) | 20 (ranking based on $\phi$) | 7 (Pareto front) |
| *R* score | 0.483 | 0.785 | 0.785 |
| *V* score | 0.387 | 0.517 | 0.517 |
| *E* score | 0.749 | 0.824 | 0.824 |
| φ mean | 0.4898 | 0.721 | 0.721 |
| Expert utility | 2.1 | 4.3 | 4.9 |
| *ETE* | 0.355 | 0.567 | 0.901 |

Relative to the *Baseline* configuration, the *Intermediate* variant improves the quality of the retained evidence, increasing the mean relevance score from 0.483 to 0.785 and the expert utility score from 2.1 to 4.3. These gains are primarily attributable to semantic filtering, evidence-level weighting, claim deduplication and contradiction consolidation. The full SCEPTER variant increased expert utility to 4.9 by replacing top-$K$ ranking with Pareto-optimal claim selection. Although the Pareto front retained only 7 claims on average compared with 20 claims in the *Intermediate* variant, the selected claim entropy increased from 0.567 to 0.901, indicating greater diversity in the retained evidence. Therefore, a smaller but more diverse evidence set produced recommendations that are judged more useful by the expert evaluators.

## 5. Conclusions

SCEPTER framework is proposed as an evidence-driven decision-support pipeline that transforms free-text clinical case descriptions into traceable, evidence-based recommendations. The framework combines PubMed retrieval, PubMedBERT semantic ranking, LLM-based claim extraction, Pareto-optimal claim selection, grounded recommendation generation and Paper Q&A interactions.

The experimental evaluation conducted on 150 case studies assessed the framework's ability to condense large volumes of biomedical literature into actionable clinical insights. On average, an initial search space of 576 PubMed papers is reduced to 53 retained papers, 7 Pareto-optimal claims and 3 final recommendations, corresponding to an overall compression ratio of 192:1. Despite this substantial reduction, the retained evidence maintained high diversity (evidence-type entropy=0.901) and a non-trivial Pareto hypervolume (0.21), indicating that the final recommendations are derived from complementary evidence profiles rather than a narrow subset of the literature. Expert evaluation involving five psychologists confirmed the practical value of the framework, with overall utility, evidence adequacy and contradiction usefulness all receiving ratings of 4.9/5, while time-saving potential achieved the maximum score of 5.0/5.

The results provide positive answers to the research questions. Regarding *RQ1*, SCEPTER transforms large literature corpora into actionable recommendations through a multi-stage evidence compression pipeline. For *RQ2*, the integration of case relevance, methodological quality and literature support within a multi-objective optimization framework enables the identification of evidence that would not have been retained through conventional ranking approaches. The ablation study demonstrated that Pareto-based selection increases the selected-claim diversity (from 0.567 to 0.901). Concerning *RQ3*, the explicit contradiction-detection mechanism identifies and surfaces conflicting findings, providing clinicians with additional context regarding the certainty and consistency of the available evidence. *RQ4* is addressed through the provenance-aware architecture and Paper Q&A interactions, which ensured that recommendations and generated answers remain traceable to supporting claims and source publications while allowing specialists to interactively explore the literature.

**Statements and declarations**

**Ethical Approval.** Not applicable

**Consent to Participate.** Not applicable

**Consent to Publish.** Not applicable

**Authors Contributions.** Contribution to the study conception and design: SVO and AB. Material preparation, data collection and analysis were performed by SVO. The first draft of the manuscript was written by SVO and AB, the second draft by SVO, and they also commented on all versions of the manuscript. The authors read and approved the final manuscript.

**Disclosure statement/Competing Interests.** The authors report there are no competing interests to declare.

**Data availability statement.** Open data from PubMed.

**Acknowledgement and Funding**: This work was supported by a grant of the Ministry of Research, Innovation and Digitization, CNCS/CCCDI - UEFISCDI, project number COFUND-DUT-OPEN4CEC-1, within PNCDI IV. This project has been funded by UEFISCDI under the Driving Urban Transitions Partnership, which has been co-funded by the European Commission.